\documentclass{article}
\usepackage{kr}

\usepackage{times}
\usepackage{soul}
\usepackage{url}
\usepackage[hidelinks]{hyperref}
\usepackage[utf8]{inputenc}
\usepackage[small]{caption}
\usepackage{graphicx}
\usepackage{amsmath}
\usepackage{amsthm}
\usepackage{booktabs}
\usepackage{algorithm}
\usepackage{algorithmic}
\usepackage{todonotes}
\setuptodonotes{fancyline, color=blue!30}

\title{Deep Algorithmic Question Answering: Towards a Compositionally Hybrid AI for Algorithmic Reasoning}

\author{%
    Kwabena Nuamah
    \affiliations
    School of Informatics, University of Edinburgh
    \emails
    k.nuamah@ed.ac.uk    
}

\begin{document}

\maketitle

\begin{abstract}
  An important aspect of artificial intelligence (AI) is the ability to reason in a step-by-step ``algorithmic'' manner that can be inspected and verified for its correctness. This is especially important in the domain of question answering (QA).
  
  We argue that the challenge of algorithmic reasoning in QA can be effectively tackled with a ``systems'' approach to AI which features a hybrid use of symbolic and sub-symbolic methods including deep neural networks. Additionally, we argue that while neural network models with end-to-end training pipelines perform well in narrow applications such as image classification and language modelling, they cannot, on their own, successfully perform algorithmic reasoning, especially if the task spans multiple domains. We discuss a few notable exceptions and point out how they are still limited when the QA problem is widened to include other intelligence-requiring tasks. However, deep learning, and machine learning in general, do play important roles as components in the reasoning process.
  
  In this position paper, we propose an approach to algorithm reasoning for QA, \emph{Deep Algorithmic Question Answering (DAQA)}, based on three desirable properties: interpretability, generalizability, and robustness which such an AI system should posses, and conclude that they are best achieved with a combination of hybrid and compositional AI. 
  
\end{abstract}

\section{Introduction}
Algorithms form the basis of problem solving and are, therefore, critical to any attempts to emulate human-like reasoning in AI. Algorithmic reasoning, as defined in \cite{velickovic2021}, allows us to automate and engineer systems that reason. 
An interesting domain in which to apply and evaluate such AI capabilities is that of question answering, and in particular, open domain QA. 
Some of the early techniques in QA, e.g. \cite{green1961}, focused on reasoning about problems in a purely logical manner. However, recent techniques have been aimed more at the challenges of constructing the right queries to retrieve answers from knowledge bases (or sources) (KBs) (\cite{usbeck2017}, \cite{dubey2019}) as well as the construction of very large language models over a large number of documents from the web \cite{devlin2019}.

Yet, many other tasks, such as the automatic selection of KBs and relevant knowledge, choice of inference algorithms, and how to combine them, are all important to fully automate the QA process. Several of these tasks are scoped out as engineering tasks which experts perform when deploying these AI systems (see Figure \ref{fig:qa_tasks}). We argue that these scoped out tasks should be part of the AI models which are built for QA tasks, as they are key ingredients in the full automation of the QA process. \emph{Deep Algorithmic Question Answering} focuses on these tasks as well as the traditional QA problem with the added task of tackling questions that require multiple steps of reasoning to solve.

In this position paper we will focus on the QA problem, especially since QA is one of the longest standing applications of AI and since many other AI problems can be framed as QA problems.
Why is it important to take such a high level perspective of QA and algorithmic reasoning, instead of one at the deeper level of knowledge/vector representation and semantics? 

\begin{itemize}
    \item It puts into sharper focus how narrow many popular AI techniques are: e.g, language modelling, image classification, etc. `Narrow' in the sense that the models excel at perception tasks for which lots of training data is available, and in the sense that they are restricted to those specific tasks and cannot be applied to other tasks `as-is' without major changes to the models or how they used.
    
    \item It highlights how unreasonable many of the assumptions in AI models are when applied to real-world problems. For instance, (1) the assumptions that the answer to a question will be from the same distribution of data which was used to train the model; (2) the assumption that the model always has access to all the data that it needs to answer a question such that the decision of choosing which KBs to use to answer a question is never a problem. In many real-world applications, the data sources are diverse and heterogeneous, noisy and incomplete.
    
    \item It shows how many AI techniques fail to address some of the challenging problems that have to be tackled. For example, dealing with uncertainty, noisy and incomplete information from KBs, especially in the context of QA.
    
    \item It shows how huge aspects of what we currently claim to be AI are heavily dependent on designs and inputs from humans and how much work needs to be done to solve simple tasks without human intervention. For instance, pre-defining which DL models are used to tackle a classification or prediction task. Although tasks such as feature engineering, which was predominantly an expert's task have been replaced by better DL models, the human expertise has only shifted to tasks related to the choice of architecture of the neural network model, dataset selection and pre-processing for training, and the general engineering required to solve the specific task at hand.
    
    \item It shows why a compositional and hybrid approach is needed given that many of the tasks cannot simply be handled with an end-to-end training of deep neural network models. We believe that a \emph{systems approach} to AI is needed to tackle algorithmic reasoning in QA and agree with the claim that there is the need to find new ways to synthesize AI from a hybrid of symbolic methods and deep neural networks \cite{marcus2019}.
\end{itemize}

We conclude that it is important to refine the scope of problems that AI for QA should solve by incorporating those tasks which, in the real-world applications, look messy and are often tackled by humans experts or data annotators. Further, tackling these problems highlights the need for AI approaches that can appropriately leverage both symbolic and sub-symbolic AI methods and also brings to the fore the need to have AI systems that are compositional in order to adapt seamlessly to different problem types.

In the sections that follow, we give some background to algorithmic reasoning, hybrid AI and compositionality, and then describe our proposed DAQA system.

\section{Background}
\subsection{Algorithmic Reasoning}
The task of algorithmic reasoning places emphasis on automating systems to reason about problems and programs following similar mechanisms to those which humans use when solving problems \cite{kroger1977lar}. However, our interpretation of this task goes beyond the classic logical reasoning context to one where learning and reasoning are combined to tackle more complex and diverse problems.
This includes, for instance, choosing which algorithms to use, when and how to combine them \cite{velickovic2021}.
Its application to question answering means having an automated system which is deliberate in the selection of inference steps needed to answer a question such that the inference process forms a computational graph which represents an algorithm for solving the problem.

We claim that achieving this with a purely symbolic or DL approach is not practical, given the known limitations of symbolic methods and sub-symbolic methods \cite{marcus2019}. 
There is a lot of work ongoing to reconcile these techniques (see section \ref{hybrid_ai}).
However, many of these are either theory-focused, or at levels of abstraction that still makes it hard to tackle algorithmic reasoning in a practical problem domain such as QA.

This work is motivated in part by models proposed in \cite{velickovic2021}. However, we note that the authors make assumptions about the contexts in which algorithms are used, and limit the concept of algorithms to a narrow application of a single algorithm that is trained end-to-end. This paper extends that notion to include the automatic, appropriate and effective composition of algorithms to solve different kinds of problems in QA. Other related work such as \cite{fader2014},\cite{liang2013}, \cite{nuamah2016functional}, and \cite{bundy2018automated} focus on the application of rules for decomposing problems in order to find answers, leading to inference processes or plans which are constructed dynamically. We extend some of these ideas in this work.


Implicitly, the expectation of algorithmic reasoning is that the process and the inferred answer can be inspected to verify the steps involved in answering a question. This is very different from the expectations one has when using deep neural network models to train end-to-end models where the process of completing the task is not interpretable. 

\begin{figure}[t]
    \begin{center}
    \includegraphics[width=1 \linewidth]{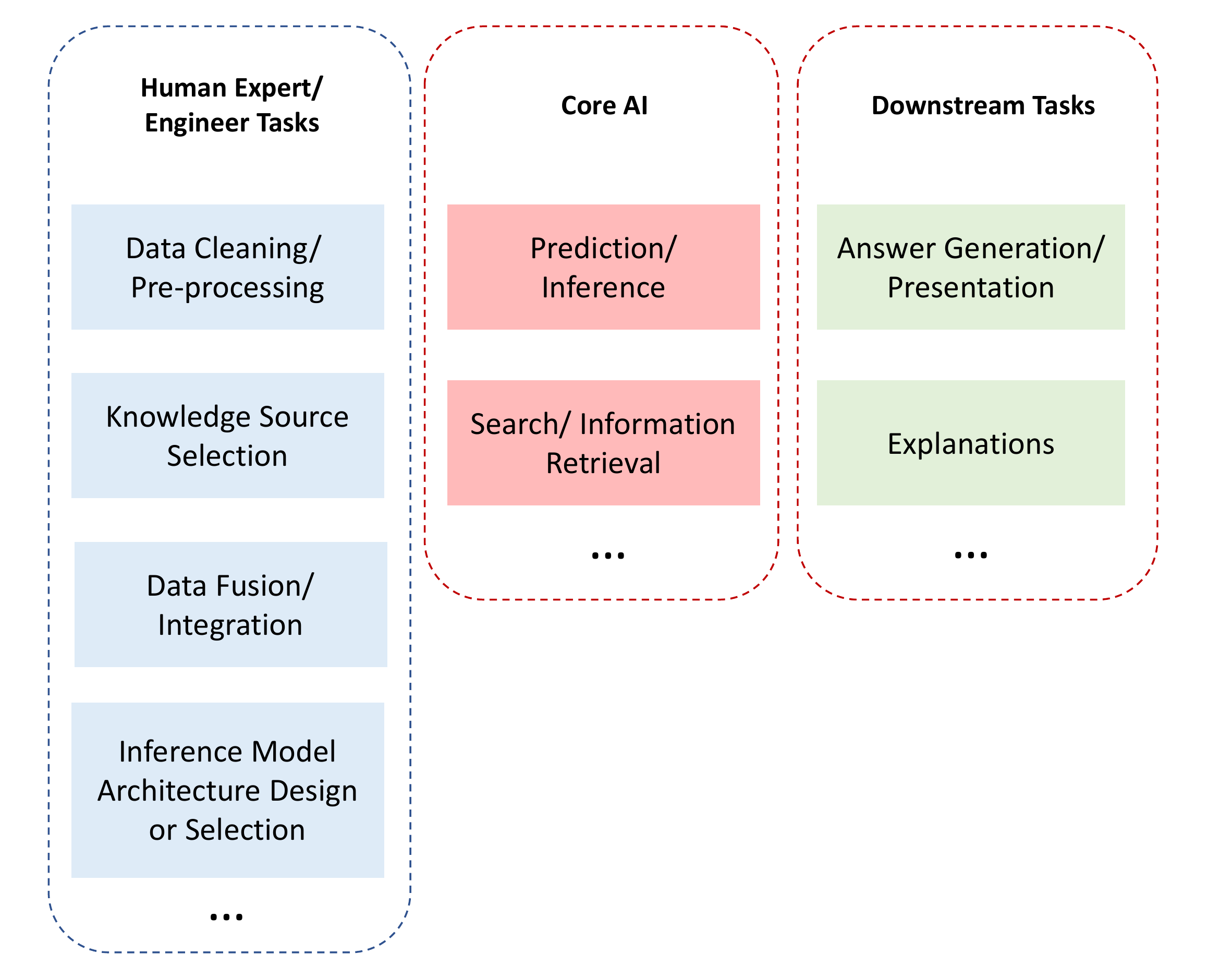}
    \end{center}
    \begin{quote}
    {\em }
    \end{quote}
    \caption{Diverse tasks that are part of the open-domain question answering process. However, most of the attention in work related to QA focus on the core AI tasks related to information retrieval, inference or prediction}
    \label{fig:qa_tasks}
    \end{figure}

\subsection{Hybrid AI}
\label{hybrid_ai}
Hybrid AI is concerned with the integration of symbolic (logical) and sub-symbolic (DL-based) AI methodologies into neuro-symbolic architectures. 
This is a rapidly growing field with diverse approaches being explored. We refer you to papers that survey these works including \cite{besold2017} and \cite{belle2020}. Additionally, Henry Kautz's classification of the different types of neural-symbolic systems integrations are outlined in \cite{lamb2020}. Our notion of hybrid AI is primarily inspired by DARPA's \emph{`Third Wave of AI'} research focus \cite{darpa2018}, ``where systems are capable of acquiring new knowledge through generative contextual and explanatory models''.

The strengths and shortcomings of both DL and symbolic AI paradigms are well documented. More recently in \cite{bengio2021}, some of the pioneers and advocates of DL for AI highlighted the need to address the limitations of DL in order to tackle some of the human-like reasoning capabilities. In particular, they mention DL's current inability to perform deliberate systematic reasoning and planning as described by Kahneman's \emph{`System 2'} reasoning \cite{kahneman2012}.

Reconciling methodologies in distinct areas of learning and reasoning (e.g. statistics and logic) means combining the respective advantages while circumventing the shortcomings and limitations \cite{besold2017}.
Some of the approaches taken to reconcile symbolic and sub-symbolic reasoning include the following (not in any way exhaustive):
creating a one-to-one correspondence between artificial neurons and elements of logical formulae \cite{riegel2020};
using reinforcement learning with Monte-Carlo tree search to play Go in AlphaGo \cite{silver2016}; differentiable architecture search by applying reinforcement learning over a discrete and non-differentiable search space\cite{liu2018darts};
combining deductive and inductive reasoning methods  for question answering \cite{nuamah2018} \cite{bundy2018automated},\cite{nuamah2020explainable};
neural networks with external memory in the Neural Turing Machine (NTM) \cite{graves2014} and reinforcement learning NTM variants \cite{zaremba2016} to make them more expressive;
memory networks \cite{bordes2015}\cite{weston2015}; probabilistic reasoning and program induction \cite{manhaeve2018}. There is also a lot of interest in \cite{cozman2021a} for enhancing machine learning with knowledge representation and reasoning. In addition, relational reasoning using neural networks is showing a lot of promise in the visual QA context \cite{santoro2017simple} \cite{santoro2018relational}. A review in \cite{aditya2019integrating} also discusses different techniques for integrating knowledge and reasoning for image understanding.

A common theme in most of the work exploring hybrid AI is the need for symbol manipulation on models of the world, while being able to leverage other sub-symbolic machinery to learn these models from examples or to predict actions based on the models. These capabilities are also essential for performing algorithm reasoning.

\subsection{Compositionality}
\label{compositional_ai}
The space of algorithms and algorithmic reasoning is far too large and varied to construct a single neural network model to solve it in practical way.
It is not always possible to program or train one AI system to solve diverse kinds of problems. In many cases, even the ability of an expert to engineer a system to solve a range of problems, such as that of open-domain QA, is limited by the fact that one cannot anticipate all the possible kinds of questions to answer and how to combine existing AI modules achieve it.  Compositionality provides a mechanism to compose solutions to problems by automating the combining of existing AI modules to solve new and varied problems.
In this work, we use ``compositionality'' in a loose sense to include the entire spectrum from the high level integration of distinct AI components and systems, through to automatic program composition, all the way to the deeper level integration of knowledge representation, semantics and neural embedding.

Different approaches can be used to build such compositional AI systems. We highlight a few below, but it is in no way an exhaustive list. \cite{gaunt2017} created an end-to-end trainable system, \textit{NEURAL TERPRET}, that learns to write interpretable algorithms with perceptual components, while the Neural Turing Machine \cite{graves2014} extend neural networks with an external memory such that the network can infer simple programs such as copying and sorting.  
Some neural-symbolic methods provide compositionality by treating the symbolic and neural network modules both as black boxes and integrating them by exposing appropriate functions \cite{tsamoura2021}.
In a majority of cases, compositionality is achieved by mapping neural network modules onto the semantic parse tree of a natural language question or the generation of sequences of functions from the question text using a trained network \cite{andreas2016} \cite{yi2019} \cite{liang2017}\cite{kapanipathi2020}\cite{johnson2017}\cite{johnson2017a}. 
The generated program is then executed to answer the question.

However, generating a neural network architecture from a semantic parse tree of natural language text is not enough to achieve algorithmic reasoning. This is because intermediate reasoning steps such has handling failure due to the lack of relevant data cannot be recovered from in a shallow parse tree without any further reasoning or inference steps. Sometimes, the data retrieved at one step during inference determines how the rest of the algorithm is developed. For instance, in a question such as ``Which country in Europe will have the highest GDP growth rate by 2032", the kind of data retrieved (or the lack thereof) will determine if retrieval is sufficient, or a more involving regression on past data for prediction will be needed. Hence, the automatic formulation of new algorithms using existing components requires one to look beyond the initial parse tree of the question and to work within the constraints of pre- and post-conditions of the underlying symbolic and sub-symbolic modules in order to combine them appropriately.


\begin{figure}[t]
    \begin{center}
    \includegraphics[width=0.6 \linewidth]{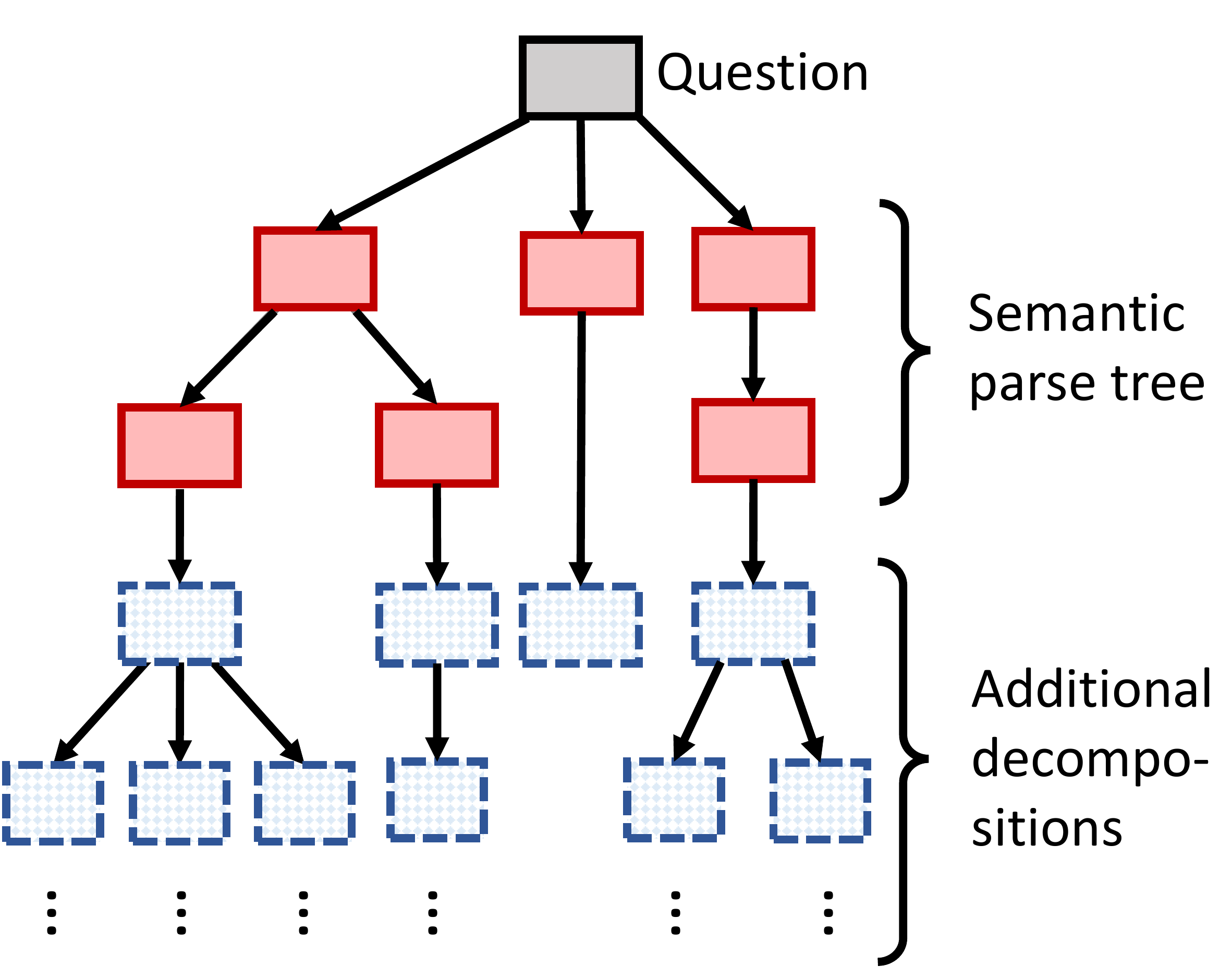}
    \end{center}
    \begin{quote}
    {\em }
    \end{quote}
    \caption{Going beyond the semantic parse tree of the question by applying additional decompositions based on rules or pre-trained models for predicting continuations of the inference plan. }
    \label{fig:decomps}
    \end{figure}

\section{Desiderata}
\label{desiderata}
Achieving the task of algorithmic reasoning in the domain of question answering requires us to have some expectation of what such a system should look like and how it should behave.  We list three of these below, all of which introduce new challenges that, if solved, will advance the development of AI architectures for QA.

\begin{itemize}
    \item \textbf{Interpretability: } One of the basic requirements of a QA system with algorithmic reasoning capabilities is that its inner workings are interpretable and inspectable by a human user. Additionally, the intepretability allows a user to check if the pre- and post-conditions of the algorithms are satisfied. For instance, if heterogeneous modules are automatically composed to form novel algorithms which answer a question, one should be able to verify that the conditions associated with the appropriate use of the modules have been met.
    A key requirement for interpretability is a representation of the inference mechanism which supports both symbolic and sub-symbolic inference. For example, a dual (or hybrid) representation which supports both deductive inference through symbol manipulation and inductive inference over data observations using statistical methods will be needed. Better still, a representation which allows for a fluid translation between these representations will be useful.
    
    \item \textbf{Generalizability: } It is also important to think of QA problems at a much broader level beyond the narrow vertical perspectives, such as image recognition, prediction or classification tasks, in order to build the capabilities of the AI systems for algorithmic reasoning. 
    One of the criticisms of narrow AI is the fact that they solve very specific problems well, but rarely capture most of the complexities which need to be dealt with in real-world applications. Most of these complexities are often handled by an expert. A desirable feature of AI systems in QA which performs algorithmic reasoning is that they are not restricted to neatly defined problems in benchmark datasets which sometimes lead to over-engineering of  AI architectures that are built to exploit biases that are observable in the data set. Additionally, it is desirable for QA systems to be general in how they compose algorithms, both in the aspects of the QA processes that they use and the in kinds of problems that they can solve.
    
    \item \textbf{Robustness: } Finally, there is a need to build AI systems that are robust in the presence of noise, incomplete data and uncertainty. Robustness is also needed as knowledge changes or new knowledge is acquired.
    These are obvious problems that are faced when using AI in the real-world and so working only on problems or data sets that exclude these challenges results in AI systems which are brittle. In algorithmic reasoning in particular, it is necessary to build AI systems which are able to identify these uncertainties and incorporate them in the inference process and in the automatic generation of programs to solve problems. For instance, failures to access data or inconsistencies in data retrieved from KBs should not stop the QA system from finding answers to questions if alternative strategies for solving the question can be found using a different algorithm. However, how the AI system deals with such issues should be transparent to users.
\end{itemize}

In summary, many of the debates about symbolic versus sub-symbolic AI cease to exist when the scope of the problem being solved is viewed in its entirety; i.e. to include not only the specific task of prediction or classification, but other intermediate reasoning and decision steps (see Figure \ref{fig:qa_tasks}) which are often performed by the creators of the AI system and left out of the scope of what the system does.
    
\section{Deep Algorithmic QA: Hybrid + Compositionality}
\label{proposal}
Our proposed approach to algorithmic reasoning for question answering, DAQA, leverages both hybrid AI and compositionality. Specifically, we are interested not only in a narrow aspect of the question answering task, but in the often ignored aspects of the tasks usually hidden under the list of things which an engineer or expert user does. DAQA is deep in two senses: (1) the inference graphs constructed are deeper than the initial semantic parse trees of the question; (2) it uses deep neural networks as part of the inference framework.

We use the following question example to shed light on the different aspects of our proposal: 
\emph{``What will be the population of the country in Europe which is predicted to have the highest GDP in 2032?''.}

\subsection{Motivation}
First, we make no assumptions about the presence of data needed to answer the question. We only assume that the AI system has a list of different KBs than it can access. These could be web documents sources that it has crawled, publicly available knowledge graphs with interfaces for querying data (e.g. SPARQL \cite{world2013sparql} or a web-based application programming interface (API)). This means that the choice of KBs to query and the integration of data from diverse sources is not trivial. Different modalities (text, images, videos) and formalisms (unstructured text, RDF, graph, probabilistic, etc.) make the task all the more difficult.

Second, we do not assume that the answer is pre-stored in any KB. In the question above, chances of having an exact answer stored some knowledge is very low to non-existent. As such, the only way to solve this question is to reason about it and dynamically construct an algorithm that can solve it.

Third, we claim that creating a deep neural network model which is trained in an end-to-end way to tackle open-domain QA including questions of the kind that we have above is not practical with the present state of the technology. However, simpler neural network models are available for solving aspects of the problem, such as the semantic parsing task and the prediction task. This brings to the fore a need for a compositional approach. That is, general purpose neural network models, statistical and arithmetic inference operations can be composed in a dynamic way to construct an appropriate algorithm that solves the question. Constraints on the individual inference modules such as pre-conditions and post-conditions ensure that they are composed in a computationally valid way.

Fourth, we do not assume that a correct semantic parse of the question is enough to compose a program which answers the question. In addition to semantic parsing, it is necessary to reason about the question to explore possible algorithms which could solve it (see Figure \ref{fig:decomps}). In the above question, for example, there are tasks such as prediction that will not be explicit in the parse tree. As such, it is important to consider deductive methods to decompose the problem. Additionally, such decomposition needs to be recursive and be robust in the event of a failure to infer an answer by exploring different possible deductions simultaneously. Hybrid AI plays a significant role here as it provide a substrate on which to perform reasoning in the inference process while offering a more rigorous inductive mechanisms to draw inferences from data.

\begin{figure}[t]
    \begin{center}
    \includegraphics[width=1 \linewidth]{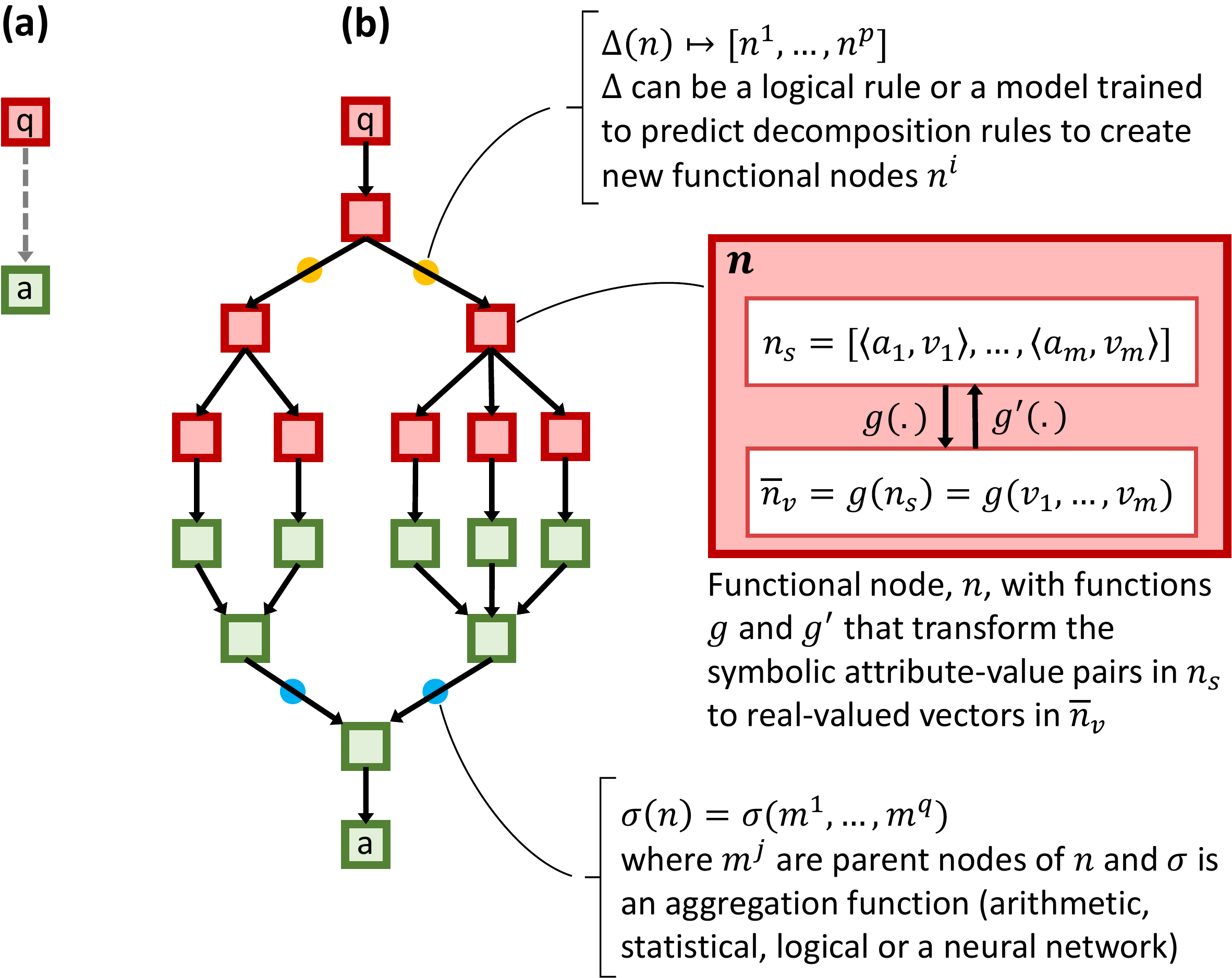}
    \end{center}
    \begin{quote}
    {\em }
    \end{quote}
    \caption{
    (a) Shows the base inference graph with a question node and an answer node that is to be inferred. They are linked by an edge that can be split by applying decomposition operations on the question node.
    (b) An inference graph made up of functional nodes and edges labelled by operations for predicting decomposition and aggregation functions. Decomposition sub-graph (in red) is guided by a function $\Delta$ that decomposes a functional node to create new continuations of the inference graph, and aggregation sub-graph (in green) which uses a model $\sigma$ to select appropriate functions to combine nodes. Functional nodes provide both a symbolic and vector representation of the node's attribute-value internal representation, as well as function $g$ and $g'$ for converting between the two representations.
    }
    \label{fig:inference-graph}
    \end{figure}
    
\subsection{Proposed Model}
A fundamental part of the above motivation is that of knowledge representation which supports both hybrid AI and compositionality. 
Although we leverage symbolic AI methods, we do not propose a classic expert system-styled mechanism. 
Instead, we propose the idea of \emph{hybrid inference graphs} with \emph{functional nodes}  and illustrate these in Figure \ref{fig:inference-graph}. An inference graph is constructed and expanded dynamically through the decompositions of its functional nodes using rules that are learned. Functional nodes represent three things:
\begin{enumerate}
    \item \emph{data}: includes parsed information from the question, data to be inferred (represented by variables to be inferred) or data retrieved from KBs or inferred and propagated from other functional nodes.
    \item the \emph{functional operations} to be applied, e.g. regression. These operation could themselves be neural networks for prediction, classification, etc.
    \item a model to convert between the symbolic and vectorized representation of the functional node, possibly obtained through an aggregation of the embeddings of its elements. 
\end{enumerate}
Functional nodes, therefore, provide support for both the symbolic manipulation of objects and the vector representation which can leverage the capabilities of DL.
The edges linking functional nodes in the graph represent rules or transition functions from the state of the functional node to the next. This provides a mechanism for decomposing functional nodes, thereby expanding the frontier of the inference graph. The rules can be provided or learned from data, allowing the inference graph as a whole to be learned. Techniques developed in reinforcement learning can be used to learn these transitions functions in order to predict subsequent decompositions of nodes on the inference graph from a handful of rules and the pre- and post-conditions of the various inference operations.


Training this system as a whole to answer questions can be achieved in two ways. First, one may use some form of weak (or distant) supervision signal such as the question and the expected answer. However, constructing such as large dataset is an expensive and prohibitive process. An alternative is to leverage existing datasets to train the individual modules and learn a model that complements the deduction process by predicting candidate decompositions to be applied and the choice of appropriate operation for aggregating functional nodes.

As new knowledge becomes available, the different sub-models needed to construct the inference graph, e.g. the functions $\Delta$ and $\sigma$, can updated without having to re-train the entire system. Also, as prior knowledge changes, the representation in the functional nodes $n$ can be updated. Similar to the method used in \cite{manhaeve2018}, uncertainty values can be inferred and stored in one of the $n_s$ attribute-value pairs. This can be the basis of Bayesian updates as prior knowledge from KBs changes.

\section{Discussion}
Our proposed approach brings on board novel perspectives on AI for question answering. However, it also builds on some other related ideas and methodologies.

While there have been QA techniques that perform deductive reasoning during inference (e.g. \cite{fader2014}) using operations such as query decomposition and rewriting, they lack the machinery to perform inductive reasoning using more detailed arithmetic and statistical operations. Recent methods such as the FRANK system \cite{nuamah2016functional}, \cite{bundy2018automated}) in the FRANK QA system adopt a hybrid inference architecture which allows for deductive reasoning using rules in a recursive manner and aggregation of data for prediction using a variety of inference operations including pre-trained neural network models. However, this approach lacks (1) a neural representation of inference nodes and (2) the ability to intelligently search through the space of inference operations for the appropriate ones to use in order to make inference more efficient. Recent work attempts to improve the automatic selection of kernels for Gaussian Process regression \cite{fletcher2021}. That said, the recursive approach used allows for the dynamic composition of modular inference operations beyond the one constructed from the syntactic or semantic parse of the question. 

AutoML (see survey in \cite{he2021}) and AutoAI \cite{wang2020} aim to automate the pipeline of ML tasks. Although they address some pipeline tasks such as data cleaning, feature engineering, model selection, etc, current methods do not address some of the reasoning tasks involved in dynamically decomposition the problem to find appropriate answers when specific inference paths fail to yield results.

Many of the QA methods discussed in \S\ref{hybrid_ai} and \S\ref{compositional_ai} generate programs based on the parse trees from the natural language question, and do not perform any further deductive reasoning or decompositions. As discussed in the respective sections, they are focused on other neuro-symbolic tasks such as integrating knowledge into neural networks and do tackle many of the tasks discussed in \S\ref{proposal}. 

Although \cite{velickovic2021} proposes ideas for achieving neural algorithmic reasoning, it differs from our proposal in two main ways. First, the notion of algorithms is at a different level of granularity. The focus in that paper is on `lower' level algorithms such as sorting. More complex algorithms which involve higher level operations such as regression for prediction and many other arithmetic and statistical operations are not explored. Secondly, estimating the outputs of the algorithm using a purely neural network approach still suffers from a lack of interpretability given that it is still a black-box from the perspective of a user. This makes it very hard to verify that the neural network is executing the algorithms correctly. 

It is worth noting the impact that large language models like GPT-3 \cite{brown2020language} have had on QA and AI in general. Despite its limitations, as discussed in \cite{marcus2020}, its role as a pre-trained model in fine-tuning tasks in other models, for example in CLIP \cite{radford2021learning}, is relevant to our proposed inference model. On its own, though, such large language models do not adequately address the desired properties we aim for.

Nevertheless, our proposed model also has some difficulties that need to be overcome. First, constructing a model which allows for the seamless conversion between symbolic and vector representations of the functional nodes across multiple domains is a hard problem and is still an active research area in neuro-symbolic AI. The space of decomposition and aggregation operations is also very large, so, appropriate search optimisations and heuristics will have to be developed to make it tractable. Promising work in architecture search using reinforcement learning \cite{liu2018darts} could offer a viable solution. Finally, training the model as a whole will be very hard, but reusing and fine-tuning pre-trained models in a plug-and-play manner within the inference architecture may be a possible solution.

\section{Conclusion}
The problem of algorithmic reasoning is one that fits well with the domain of QA since it helps to automate several aspects of the QA pipeline and leads to interpretable models for answering questions. We claim that a hybrid approach to AI with a strong element of compositionality is needed to tackle such a perspective on QA and other AI problems. We have proposed a systems approach to AI which leverages both symbolic and sub-symbolic methods in a framework that leads to solutions which are not possible by either one of these paradigms alone.

\section*{Acknowledgment}

The author would like to thank Vaishak Belle, Alan Bundy and Thomas Fletcher for feedback on an earlier draft and Huawei for supporting the research on which this paper was based under grant HO2017050001B8s. The author would also like to thank reviewers for valuable feedback.

\bibliographystyle{kr}
\bibliography{deep_algorithmic_qa}

\end{document}